\documentclass[a4paper, 10 pt, conference]{hsmr}
\pdfoutput=1
\usepackage[utf8]{inputenc}
\IEEEoverridecommandlockouts                       
\usepackage{mathtools}
\usepackage{geometry}
\usepackage[labelsep=space]{caption}
\usepackage{newtxmath,newtxtext}
\usepackage{authblk}
\usepackage{pgfplots}
\usepackage{graphicx}
\usepackage[colorlinks,urlcolor=blue]{hyperref} 
\usepackage{newtxtext,newtxmath}

\pgfplotsset{compat=newest} 
 
\title{\LARGE \bf
Using LOR Syringe Probes as a Method to Reduce Errors in Epidural Analgesia -- a Robotic Simulation Study} 

\author{\LARGE N. Davidor$^{1}$}
\author{\LARGE Y. Binyamin$^{2}$} 
\author{\LARGE T. Hayuni$^{2}$} 
\author{\LARGE I. Nisky$^{1}$} 

\affil{\Large\textit{$^{1}$Department of Biomedical Engineering, Ben-Gurion University of the Negev}\\ \Large\textit{$^{2}$Department of Anesthesiology, Soroka Medical Center}\\ \Large\textit{nitsanti@post.bgu.ac.il}}

\begin{document}

\maketitle
\thispagestyle{empty}
\pagestyle{empty}

\section*{INTRODUCTION}
In epidural analgesia, anesthetics are injected into the epidural space, to block signals from traveling through nerve fibres in the spinal cord or near it. To do so, the anesthesiologist inserts a Touhy needle into the patient's skin and uses it to proceed to the epidural space, while using the haptic feedback received from a "loss of resistance" (LOR) syringe to sense the environment stiffness and identify loss of resistance from potential spaces. The two most common errors or complications of epidural analgesia are failed epidurals (FE) -- halting the needle insertion in a superficial location, which will cause no pain relief -- and dural punctures (DP), leading in most cases to post dural puncture headache (PDPH). 

The task of identifying the epidural space correctly and stopping the needle insertion while in it is challenging mechanically, and requires extensive training \cite{konrad_learning_1998}. Hence, robotic simulation is an attractive method to help optimize skill acquisition \cite{davidor}. Another advantage of robotic simulation is the ability to record kinematic information throughout the procedure, to evaluate users' performance and strategy. In this study, we used a bimanual robotic simulator that we developed in previous work \cite{davidor} to analyze the effect of LOR probing strategies on procedure outcomes.

\section*{MATERIALS AND METHODS}
\begin{figure}[ht!]
\centering
\includegraphics[width=\columnwidth]{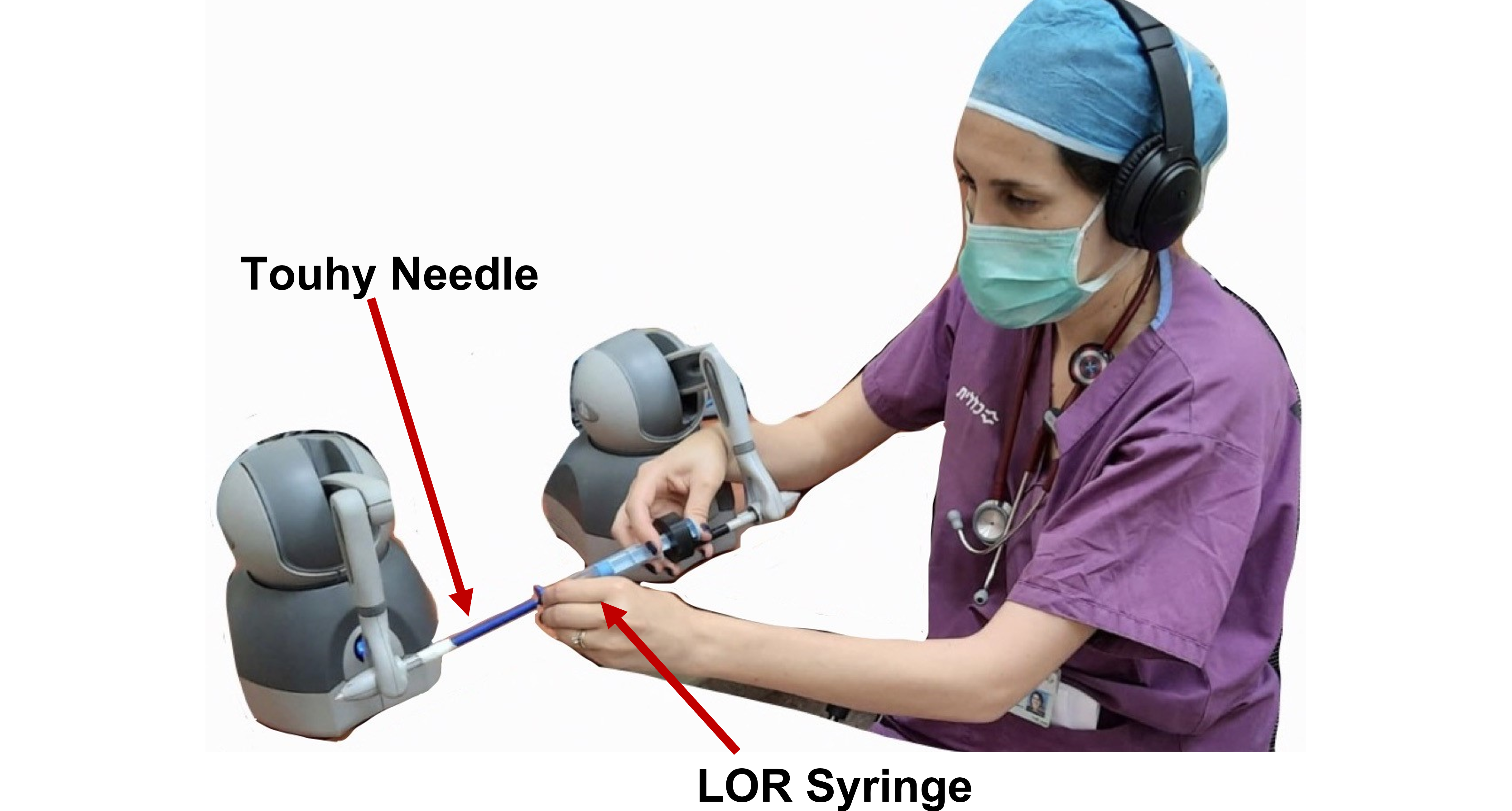}
\caption{The experimental setup: a haptic bimanual simulator for epidural analgesia. One haptic device is connected to a Touhy needle and the other is mounted by an LOR syringe.}
\label{fig_sim}
\end{figure}

Our experimental setup was comprised of a haptic bimanual simulator (Fig.~\ref{fig_sim}), that has been validated by experts \cite{davidor}. The simulator emulates the forces applied on the Touhy needle and the LOR syringe throughout the procedure (Fig.~\ref{example_trials}a), based on a force model proposed in \cite {brazil_haptic_2018}, and allows for patient weight variability and recording kinematic data. 

23 anesthesiologists of different competency levels (division into levels was based on years of experience, case count and position -- resident or attending \cite{davidor}) participated in two experiments. The first experiment ($N_1=15$) included three familiarization trials, in which there was a constant patient body mass, followed by 12 test trials that involved three different patient body masses (55, 85 and 115 kg). The second experiment ($N_2=8$) included only the test trials. To eliminate differences in familiarization between the two experiments, we used only the final nine trials of experiment two for our analyses.

To obtain LOR syringe probing movements, we subtracted the trajectory of the haptic device that was connected to the Touhy needle from the trajectory of the haptic device that was mounted by the LOR syringe. We then took the peaks of the adjusted trajectory and enumerated them. 

To examine if there were differences in probing amounts between different outcome trials, we plotted the mean number of probes performed in successful trials as a function of the mean number of probes performed in unsuccessful trials. This analysis is impossible for participants with success rates of 0\% or 100\%, and hence they were excluded from this analysis ($N_{ex.}=4$). We used the two-sided Wilcoxon signed rank test to compare between mean number of probes in successful and unsuccessful trials, since our data did not distribute normally.

To further delve into probing differences between different outcome trials, we examined the mean number of probes performed in each layer in the epidural region, in successful and unsuccessful trials. To avoid bias that is rooted in the layer thickness, we normalized the mean number of probes in each layer by dividing it by the layer thickness. This part of the study is exploratory, and hence we chose not to perform statistical analysis for testing hypotheses, and instead use this part of the analysis as preliminary investigation for a future hypothesis driven study.


\section*{RESULTS}
\begin{figure}[h!]
\centering
\includegraphics[width=\columnwidth]{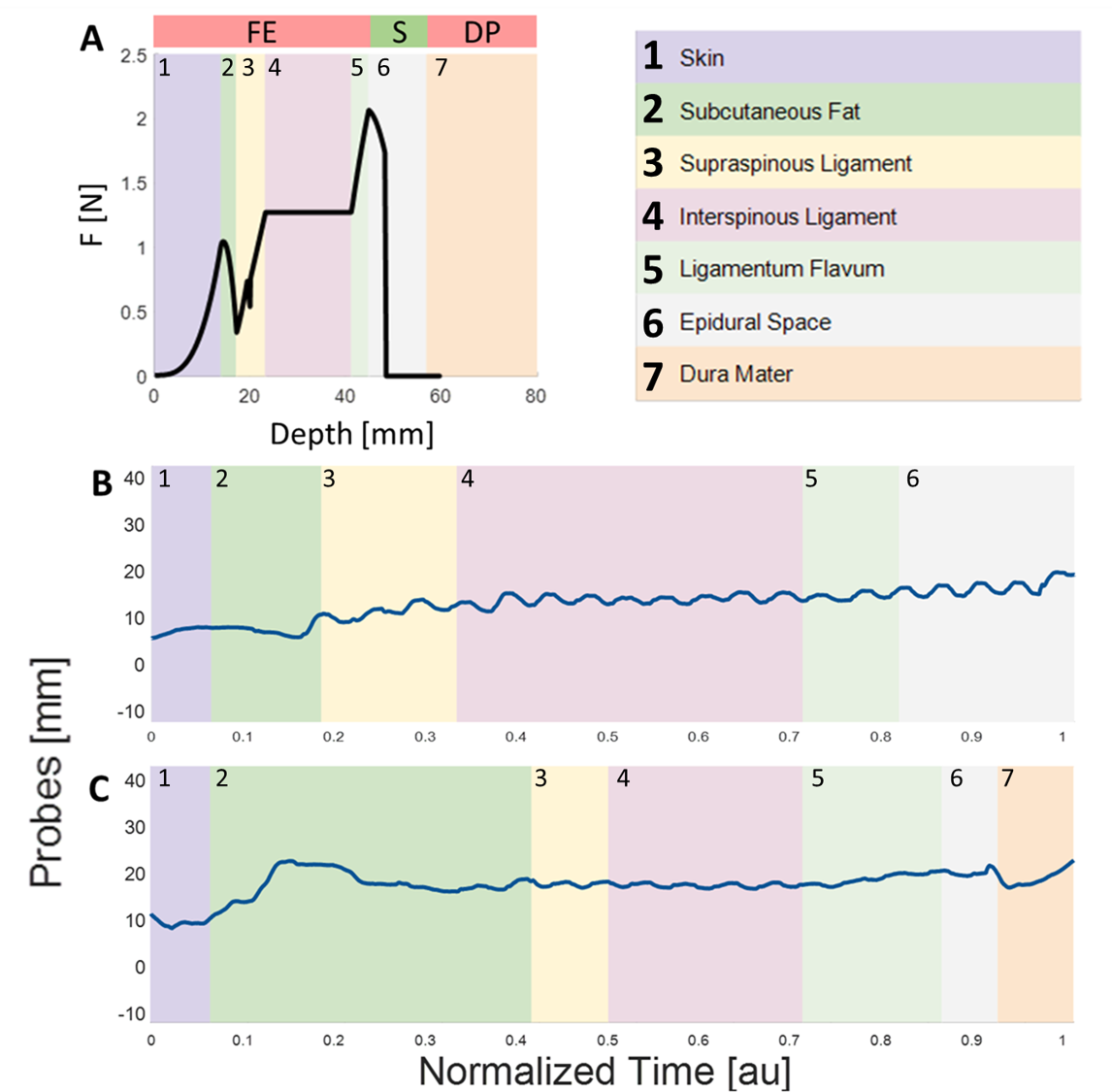}
\caption{Exerted forces and probes in two example trials. (A) Forces exerted by the haptic devices as a function of  needle insertion depth. (B)-(C) The trajectories of the LOR syringe haptic devices as a function of the normalized time in is a successful trial (B) and an unsuccessful trial (C). The different background colors and numbering in all panels represent the layers in the epidural region. }
\label{example_trials}
\end{figure}

We present examples of probes in two trials (Fig. \ref{example_trials}): a successful trial, and a dural puncture. Consistently with the trend observed in most of the participants and trials, more probes were performed in the successful trial.

\begin{figure}[h!]
\centering
\includegraphics[width=7cm]{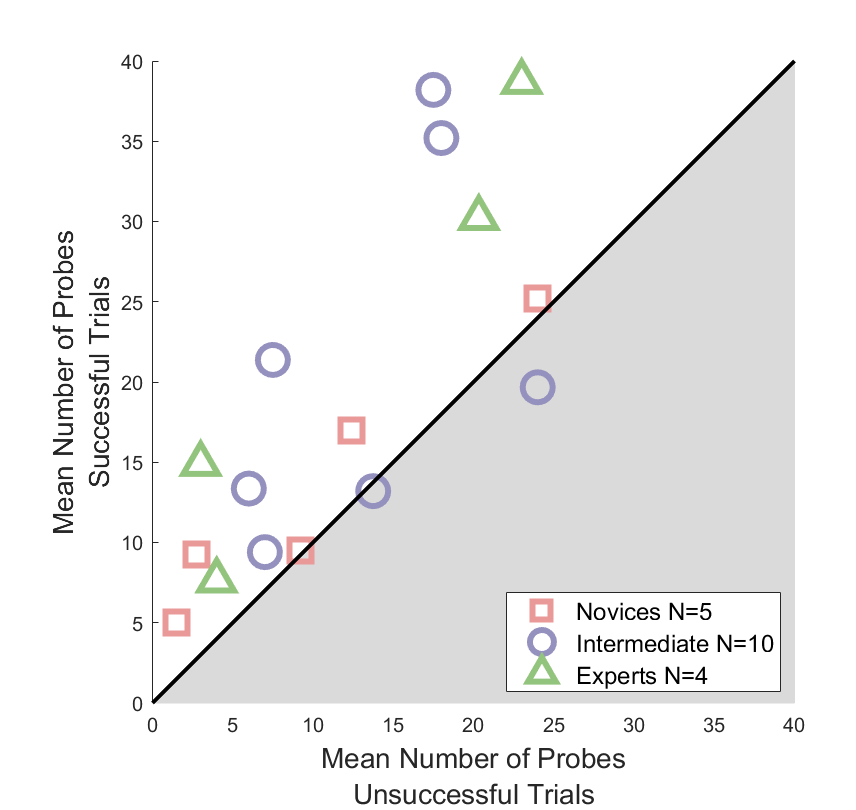}
\caption{The mean number of probes observed in unsuccessful trials as a function of the mean number of probes observed in successful trials. Each symbol represents one participant, and the different marker types and colors refer to the participant level.}
\label{XY_analysis}
\end{figure}

Studying the number of probes performed in successful trials against umber of probes in unsuccessful trials (Fig. \ref{XY_analysis}) demonstrates that most participants were above the line of equality, indicating that they performed more probes in successful trials, compared to unsuccessful trials (two-sided Wilcoxon signed rank test yielded $p=0.0018$). This result was not affected by participant level. 

\begin{figure}[ht!]
\centering
\includegraphics[width=\columnwidth]{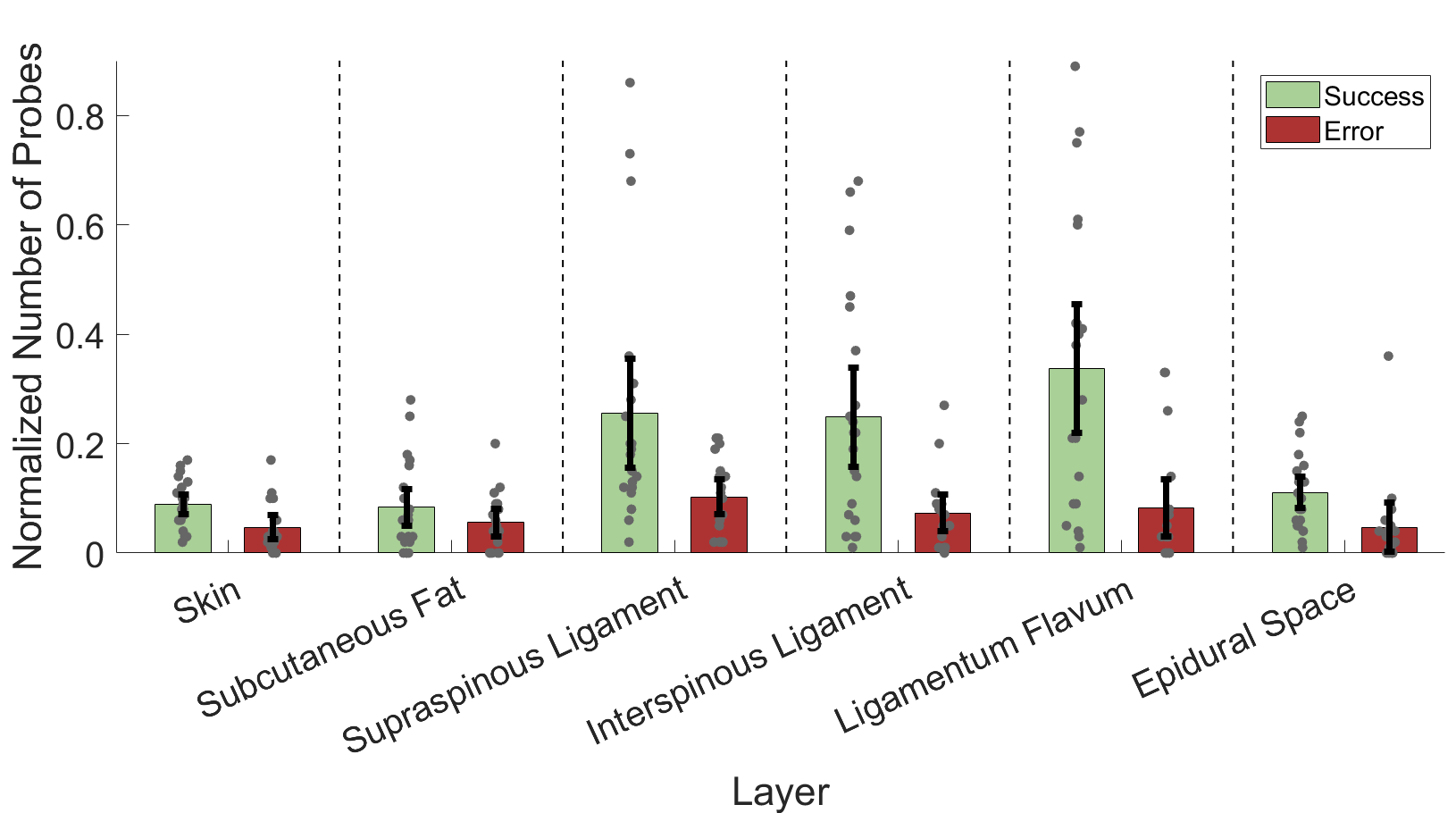}
\caption{Mean normalized number of probes per trial corresponding to layer in epidural region. Green bars represent successful trials and red bars represent unsuccessful trials. Dashed black vertical lines represent the separation between the layers. The gray points represent the mean number of probes performed by each participant in the relevant layer, and the black bars represent the non-parametric mean 95\% confidence intervals.}
\label{probes_layers}
\end{figure}

Examining the locations of probes within the epidural region (Fig. \ref{probes_layers}) revealed that more probes were performed in successful trials compared to unsuccessful trials in all layers, and the most prominent differences were observed in the three layers preceding the epidural space.


\section*{DISCUSSION}
We used a haptic bimanual simulator to evaluate the effect of probing with the LOR syringe on procedure outcomes in epidural analgesia. We found that the majority of participants probed more in successful trials compared to unsuccessful trials. Furthermore, our results suggest that this difference is more prominent in the three layers preceding the epidural space; we posit that this is caused due to higher caution when closer to the epidural space. These results indicate that a more extensive use of the LOR syringe (and more specifically, when approaching the epidural space) can assist in reducing errors in epidural analgesia. We argue that these findings may be useful in training anesthesia residents (when training in the virtual environment or the real one): instructing novices to focus on probing in relevant locations may enhance learning and produce better procedure outcomes. 

\nocite{*}

\bibliographystyle{IEEEtran}
\bibliography{hamlynref}

\end{document}